\documentclass[a4paper,fleqn]{cas-dc}

\usepackage[numbers]{natbib}
\usepackage[ruled,vlined]{algorithm2e}
\usepackage{wrapfig}

\begin{document}
\let\WriteBookmarks\relax
\def\floatpagepagefraction{1}
\def\textpagefraction{.001}

\shorttitle{Adaptive adversarial training method for improving multi-scale GAN based on generalization bound theory}
\shortauthors{J.Tang et~al. }

\title[mode = title]{Adaptive adversarial training method for improving multi-scale GAN based on generalization bound theory}

\tnotetext[1]{This  work  was  supported  in  part  by  the  National  Science  Foundation  of China under Grant 91948301, 51820105008, 51905183.}


\author[1]{Jing Tang}[style=chinese,
auid=000,
bioid=1,
orcid=0000-0002-9430-9660]
\ead{j_tang@hust.edu.cn}

\author[1]{Bo Tao}[style=chinese,
auid=000,
bioid=1,
orcid=0000-0002-4058-6128]

\ead{taobo@hust.edu.cn}

\author[1]{Zeyu Gong}[style=chinese,
auid=000,
bioid=1,
orcid=0000-0002-8276-675X]
\ead{gongzeyu@hust.edu.cn}

\author[1]{Zhouping Yin}[style=chinese]


\address[1]{The State Key Laboratory of Digital Manufacturing Equipment  and  Technology,  Department  of  Mechanical  Science  and Engineering, Huazhong University of Science and Technology, Wuhan 430074, China}

\cortext[cor1]{Corresponding author: Bo Tao and Zeyu Gong}

\begin{abstract}
In recent years, multi-scale generative adversarial networks (GANs) have been proposed to build generalized image processing models based on single sample. Constraining on the sample size, multi-scale GANs have much difficulty converging to the global optimum, which ultimately leads to limitations in their capabilities. In this paper, we pioneered the introduction of PAC-Bayes generalized bound theory into the training analysis of specific models under different adversarial training methods, which can obtain a non-vacuous upper bound on the generalization error for the specified multi-scale GAN structure. Based on the drastic changes we found of the generalization error bound under different adversarial attacks and different training states, we proposed an adaptive training method which can greatly improve the image manipulation ability of multi-scale GANs. The final experimental results show that our adaptive training method in this paper has greatly contributed to the improvement of the quality of the images generated by multi-scale GANs on several image manipulation tasks. In particular, for the image super-resolution restoration task, the multi-scale GAN model trained by the proposed method achieves a 100\% reduction in natural image quality evaluator (NIQE) and a 60\% reduction in root mean squared error (RMSE), which is better than many models trained on large-scale datasets.
\end{abstract}


\begin{highlights}
	\item The PAC-Bayes boundary theory is {\textbf{firstly}} introduced into the training analysis of multi-scale GAN, and the boundary model of multi-scale GAN is simplified using TTUR to obtain a tighter generalization bound.
	
	\item Using the tighter generalization bound, an adaptive adversarial training method is proposed that can  \textbf{substantially improve} the quality of general image manipulation task using multi-scale GAN.
		
	\item Extensive ablation experiments are conducted on different image generation tasks to verify the gain of the adaptive adversarial training approach on multi-scale GAN.
	
	\item On the image super-resolution reduction task, multi-scale GAN with adaptive adversarial training based on single image training \underline{\textbf{shows better results than GAN models trained on large-scale datasets}}. Compare to the raw multi-scale GAN model, the model trained by the proposed method achieves a 100\% reduction in natural image quality evaluator (NIQE) and a 60\% reduction in root mean squared error (RMSE).	
\end{highlights}

\maketitle

\section{Inroduction}\label{Intro}
GANs \cite{1} are models based on generators and discriminators to achieve image-to-image (I2I) transition. The GANs trained on multiple images have been used in image style transfer \cite{2,3,4,5,6}, image super-resolution restoration \cite{7,8,9,10,11,12} and many other exciting applications in the field of general image manipulation. In recent research, deep internal learning based on GANs has attracted a lot of interest as some of the proposed GANs trained on a single sample have shown comparable results to that trained on large datasets.  

In deep internal learning, features of single sample image are learned by a single model and then the model is applied to general image manipulation task. Traditional deep internal learning mostly uses a deep image prior \cite{13,14,15} approach to obtain information about sample image. However, such deep image prior approaches have some limitations as they can only operate on the image on which they are trained. If we want to manipulate a new image, the model needs to be retrained. That is, these deep image prior methods are image-specific. To address this problem, MGANs \cite{16} introduces the structure of GANs to deep internal learning. This GAN-based approach to deep internal learning does not have an image-specific requirement, which draws attention to the superiority of GANs in deep internal learning.

Although MGANs can operate on unspecified images, the model still needs to be retrained if the trained MGANs need to be applied to a different task, which means MGANs are task-specific. To solve the image-specific and task-specific problem, Shaham et al. proposed multi-scale GAN structure which can learn the features of one image and migrate the features of this image to a different domain for image manipulation task, also known as SinGAN \cite{17}. SinGAN is more efficient in achieving multiple image generation tasks as only one model is required to train for acquiring the features of the source image, and it can be directly applied to many different image generation tasks. SinGAN is undoubtedly a landmark work in introducing multi-scale GAN structures to the field of deep internal learning, and inspired followed work. The follow-up SinIR \cite{18} improves the model based on the basic idea of SinGAN, which enables the training of multi-scale GANs to be nearly 20 times faster. Hinz et al. also proposes ConSinGAN \cite{19} to accelerate the training of multi-scale GANs. It is worth noticing that these works after SinGAN have more focused on the acceleration of multi-scale GANs. In fact, the biggest problem with such multi-scale GANs limited by single image example is the lack of generated images quality.

For this reason, this paper focuses specifically on the problem of generated image quality, and introduces adversarial training approach to improve the multi-scale GANs model. We first use Two Time-scale Update Rule (TTUR) \cite{23} in the training of multi-scale GANs, which is an important trick that not only improves the performance of the GANs but also allows the iterations of the discriminator and generator to reach Nash equilibrium throughout the training process. Then we utilize Nash equilibrium to analyse the variation of model generalization bound \cite{20} under different adversarial attacks \cite{21,22}, and finally propose an adaptive adversarial training method which can greatly improve the ability for image manipulation of multi-scale GANs. Compared with the original multi-scale GAN \cite{17}, our method performs better on a variety of image manipulation tasks.

\section{Related Works}
To better understand the cutting-edge research and basic concepts of this paper, we present related works in detail. Firstly, we introduce the research on deep internal learning in order to better understand the application scenarios of the proposed approach. Afterwards the history and current state of research in adversarial training is introduced for better understanding the benefits of adversarial training method. Finally, we present the state-of-the-art research in generalization bound theory, which is the most important theoretical weapon of our adaptive adversarial training method.

\subsection{Deep Internal learning}

Internal learning means learning and modelling the features of single image sample to perform general image manipulation tasks with prior knowledge. Research in this area is well developed, and many scholars have used internal approaches to implement texture synthesis \cite{24}, image super-resolution restoration \cite{8,9,10,11,12}, image editing \cite{25}, image segmentation \cite{26}, and many other exciting image manipulation tasks. With the rapid development of deep learning, the concept of deep internal learning has been proposed. Ulyanov et al. \cite{13} proposed the deep image prior method to build a deep learning model based on a fixed image, and then use the trained model to complete a variety of different general image manipulation tasks. DCIL \cite{15} continued this idea by introducing the spatial information between pixels and constructing an image restoration model based on the basic idea of DIP, which achieved better results. Pan et al. \cite{27} conducted a more in-depth study of model optimization for image restoration based on DCIL. Gandelsman et al. \cite{14} from the same time extended the idea of DIP. They proposed a method for deep image prior based on a pair of images, implementing the target and retarget tasks for images. Shocher et al. \cite{28} proposed a zero-shot convolutional neural network (CNN) based image super-resolution restoration network, which is comparable to image super-resolution restoration based on large datasets while their model is only trained on a single image. All of the above studies demonstrate the feasibility of deep internal learning. However, there is a serious problem that traditional deep internal learning can only perform general image operations on the trained images, and cannot migrate the model to unseen images. This means that these works are image-specific.

MGANs \cite{16} provided a new perspective for subsequent research by introducing GANs into deep internal learning, which achieved non-image-specific texture synthesis. However, MGANs cannot implement other image manipulation tasks, which means they are task-specific. In order to address the image-specific and task-specific problems of the previous study, Shaham et al. proposed a novel and effective multi-scale GANs structure, and named their model SinGAN \cite{17}. SinGAN is a multi-scale GAN model which can perform internal learning through a single image and then accomplish different image manipulation tasks by performing different projections on different scales. The subsequent SinIR \cite{18}, ConSinGAN \cite{19}, and SinGAN-GIF \cite{29} are some of the work on multi-scale GAN for model acceleration and potential explorations for general image manipulation tasks. In contrast to the focus of their research, this paper centers on improving the image manipulation capabilities of multi-scale GANs.

\subsection{Adversarial Training}

The training problem of neural networks can actually be reduced to the real risk minimization problem as follow.


\begin{equation}
R_D(h)=\underset{(x, y) \sim D}{\mathbb{E}} \ell(h,(x, y))
\end{equation}


Where $h$ is the neural network model, $(x, y)$ is the data taken from the real dataset $D$, and $\ell(\cdot)$ is the loss function. The training objective of the network is to minimize the risk function $R_D(h)$. However, since the $D$ dataset is unknown, the risk function can actually only be estimated from an empirical dataset $S$, and $\left(x_i, y_i\right)$ is the sample data from empirical dataset $S$. Thus, what we actually get is the empirical risk minimization (ERM) function \cite{30} as follow.


\begin{equation}
R_S(h)=\frac{1}{m} \sum_{i=1}^m \ell\left(h,\left(x_i, y_i\right)\right)
\end{equation}


This allows us to perform optimization of $h$ based on empirical data $S$. Unfortunately, there is still a non-negligible gap between the real and empirical datasets. Chapelle et al. \cite{31} introduced the concept of Vicinal Risk Minimization (VRM). Specifically, VRM uses the vicinity of the training set to augment the empirical dataset $S$, which in turn is able to enlarge the support of the training distribution. Simard et al. \cite{32} demonstrated that such a data augmentation method consistently leads to improved generalization. Adversial training is a training method that injects spatially constrained perturbation into the training set during the training process to achieve data augmentation, which can effectively improve the generalization performance of the network. 


\begin{equation}
perturbation =\left\{\epsilon \in \mathbb{R}^d \mid\|\epsilon\| \leq r\right\}
\end{equation}


\begin{equation}
R_S^{\mathrm{Adv}}(h)=\frac{1}{m} \sum_{i=1}^m \max _{\epsilon \in B} \ell\left(h,\left(x_i+\epsilon, y_i\right)\right)
\end{equation}


FGSM \cite{33} is a famous classical adversarial training method, which performs noise injection by directly adding a small perturbation in the direction of the optimal gradient descent of the training dataset. FGM \cite{22}, on the other hand, scales the gradient according to the specific gradient and divides the injected noise by the L2-norm of the gradient. Since FGM directly computes the adversarial perturbation at once via the epsilon parameter, the perturbation may not be optimal. PGD \cite{21} performs multiple injections of noise and computation of new gradients in a single iteration, achieving a more targeted noise injection. Considering that PGD requires multiple gradient computations, it will affect the training speed during the actual training process. FreeAT \cite{34} reuse the gradient of the previous epoch when injecting noise, which can solve the problem of training speed to a certain extent. YOPO \cite{35} treat neural networks as dynamical systems and achieve a reduction in computation through the structure of the first few gradients, which also decreases the computation time of PGD. As mentioned earlier, this paper focuses on improving the accuracy of the model. Therefore, in order to better maintain model accuracy for adversarial training, we focus on two classical attacks, FGM and PGD.

\subsection{Generalization Bounds}

Adversarial training is focusing on the gap between the dataset fitted by the model and the real dataset on the model, which also named the generalization of the model. In recent years, several scholars have worked on constructing numerical models to quantify the generalization bound of these trained models. Neyshabur et al. \cite{36} provides generic generalization bounds for group norm and states that generalization ability and network depth have the same exponential dependence. Keskar et al. \cite{37} proposed a generalization bound estimation approach based on sharpness, which performs good under SGD optimization with different batch sizes. However, considering that sharpness is not a scale invariant measure, it cannot accurately predict the generalization behaviour.  Bartlett et al. \cite{38} proposed a method for estimating generalization bounds based on L1-norm and spectral norm. Considering the L1-norm is strongly influenced by the number of neurons, it is not applicable to the generalization bound analysis of current networks. In contrast to the above methods, Dziugaite et al. \cite{39} started from the classical PAC-Bayes generalization bound theory \cite{20,40} and analyzed the generalization bound of the model according to different parameter variations. Eventually they achieved non-vacuous generalization behavior predictions.

Since then PAC-Bayes generalization bound theory has attracted the attention of many scholars, and subsequent works have introduced PAC-Bayes theory to neural networks. Montasser et al. \cite{41} studied the generalization bound for trained neural networks in finite VC dimension. However, there is a large constant in their proposed numerical generalization bound, making it difficult to perform realistic training constraints. Farnia et al. \cite{42} performed an in-depth analysis of the generalization bounds for neural networks under different attacks based on the classical PAC-Bayes generalization bound and consistent convergence. Viallard et al. \cite{43} started from a different basic idea. They used pure PAC-Bayes analysis with majority votes to formulate the generalization performance of different models. It is worth noting that the above studies are mostly analyzed for non-specified models, and therefore more general conclusions can be drawn. However, in practical applications, the generalization bound inferred from the above studies is sometimes considered vacuous as there are still small differences in the generalization ability of models with different structures. In this paper we focus on the study of multi-scale GANs and analyze their structural properties in depth. We expect to obtain tighter upper bounds on the generalization error of such models and then use the upper bounds to guide model training.

\section{Multi-scale GAN architecture}

In this section we will briefly describe the structure of this multi-scale GAN as the subsequent theoretical analysis and experiments are based on it. The structure of the multiscale GAN we use here is consistent with which in SinGAN \cite{17}. Unlike traditional GAN models, multi-scale GAN contains multiple generators $\left\{G_0, \ldots, G_N\right\}$ and discriminators $\left\{D_0, \ldots, D_N\right\}$, which is shown in Figure 1. It can generate new graphs $\left\{x_0, \ldots, x_N\right\}$ at different scales by down sampling fake graph, after which the new fake graphs $\left\{x_0, \ldots, x_N\right\}$ and noise map at different scales are input to the corresponding generators and discriminators for training. Each generator $G_n$ is responsible for generating real image samples and generating patch distribution in the corresponding image $x_n$. Each discriminator $D_n$ is responsible for discriminating whether the image is a fake image generated by the generator or not. Here the image samples are generated starting from the coarsest scale, passing through all the generators in turn until the finest scale, and noise are injected at each scale. All the generators and discriminators have the same scale as the input image, so as the height of the pyramid network increases, more and more image details are captured.

Since the perceptual field at different levels is generally $1 / 2$ the size of the previous level, each generator $G_n$ at finer scales $(\mathrm{n}<\mathrm{N})$ generates more details. That is, in addition to the noise map, each generator $G_n$ accepts an enlarged version of the image from the coarser scale, mathematically expressed as follows. Here $\uparrow^r$ denotes up-sampling of the fake image at the $r$-th scale.


\begin{equation}
\tilde{x}_n=G_n\left(z_n,\left(\tilde{x}_{n+1}\right) \uparrow^r\right), n<N
\end{equation}


Notably, all the generators have a similar structure as shown in Figure 2. In each generator, the noise map $z_n$ is added to the image $\left(\tilde{x}_{n+1}\right) \uparrow^r$, and then fed into a series of convolutional layers, which can keep the generator from over-fitting. And convolution operation also allows the generator to have enough power to refine the missing details in $\left(\tilde{x}_{n+1}\right) \uparrow^r$. Each $G_n$ performs the operation shown below.


\begin{equation}
\tilde{x}_n=\left(\tilde{x}_{n+1}\right) \uparrow^r+\psi_n\left(z_n+\left(\tilde{x}_{n+1}\right) \uparrow^r\right)
\end{equation}


Where $\psi_n$ is a fully convolutional neural network containing five convolutional blocks in the form of Conv $(3 \times 3)$-Batch Norm-ReLU \cite{44}. Each convolutional block in the coarsest scale generator contains 32 kernels, after which the kernels are increased by a factor of two for every four down-samplings. Also, as the scale changes, the scale of the superimposed noise map changes.

\begin{figure}[!t]
	\centering
	\includegraphics[width=3.2 in]{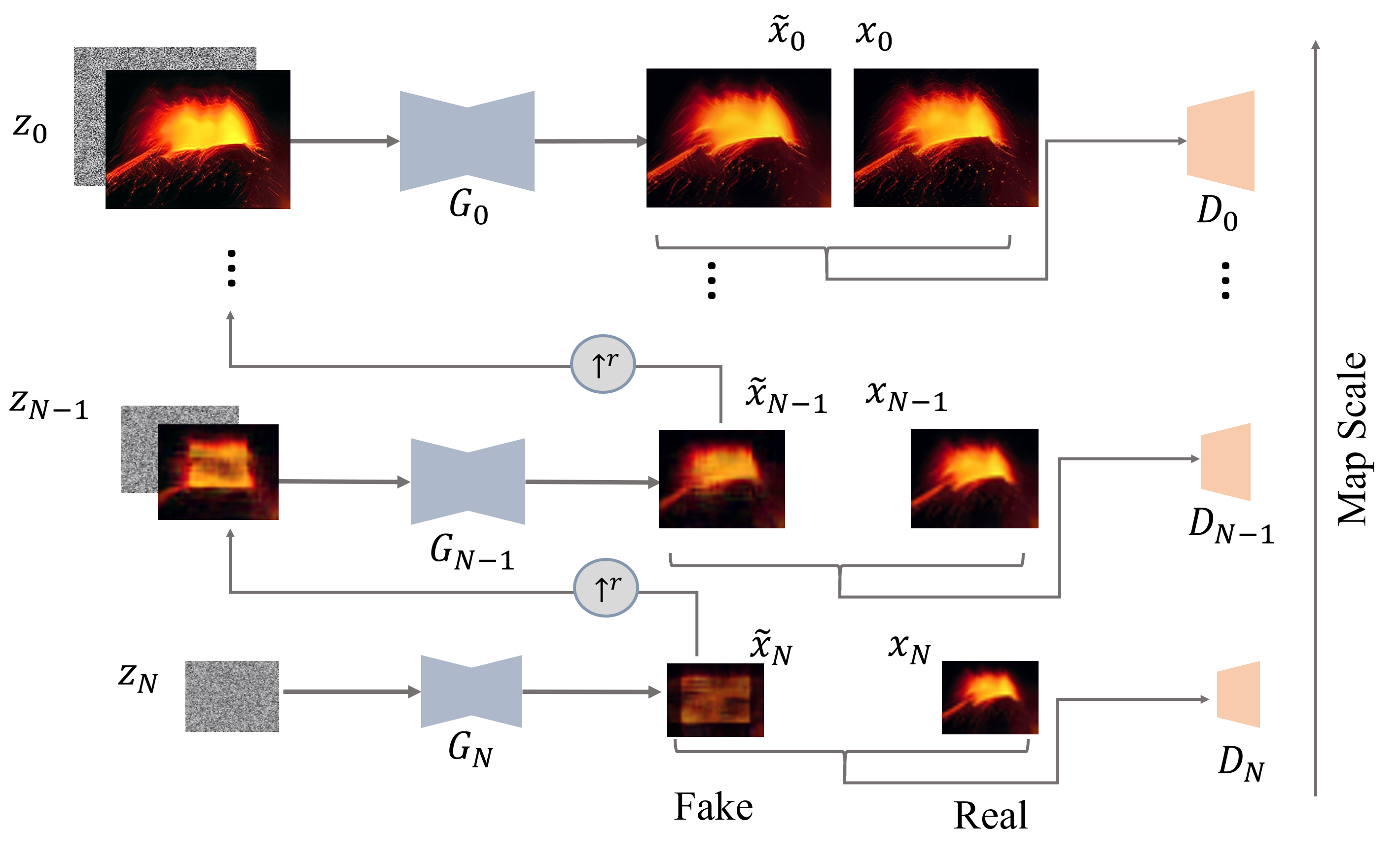}
	\caption{Architecture of Multi-scale GAN}
	\label{fig_1}
\end{figure}


\section{Adaptive Adversarial Training Method Based on Generalization Bound Theory}
In this section, we use PAC-Bayes boundary theory as a powerful weapon to propose tighter generalization error bounds for multi-scale GANs. We analyze the variation of the generalization error bounds for multi-scale GANs under different adversarial training and use this variation to guide the training of the model. Ultimately, an adaptive adversarial training method for multi-scale GANs is proposed.

\subsection{Traditional Training Method}

Traditional multi-scale GANs are trained in the order of scale coarseness as mentioned in SinGAN \cite{17}. After the generators and discriminators of the same scale are trained, the weight values of the generators and discriminators is fixed for the next scale training. The loss function in each scale is shown as Eq. 7.


\begin{equation}
\min _{G_n}\left[\max _{D_n}\left[\mathcal{L}_{\mathrm{adv}}\left(G_n, D_n\right)+\alpha \mathcal{L}_{\mathrm{rec}}\left(G_n\right)\right]\right]
\end{equation}

Where the adversarial loss $\mathcal{L}_{\text {adv }}$ intend to calculate the difference between the original image $x_n$ and the fake samples $\tilde{x}_n$ generated by the generator. In each scale of multi-scale GAN, discriminator $D_n$ will discriminate whether the corresponding patches sample from $\tilde{x}_n$ are fake images. The loss function we use is the WGAN-GP loss \cite{45}, which is consistent with SinGAN \cite{17}. The $\mathcal{L}_{\text {rec }}$ in Eq. 7 is the reconstruction loss to ensure that each generator can produce a fake image similar to $x_n$, and it is calculated as Eq. 8 .


\begin{equation}
\mathcal{L}_{\text {rec }}=\left\{\begin{array}{c}
	\left\|G_n\left(Z_n,\left(\tilde{x}_{n+1}^{\text {rec }}\right) \uparrow^r\right)-x_n\right\|^2, n<N \\
	\left\|G_N\left(Z_N\right)-x_N\right\|^2, n=N
\end{array}\right.
\end{equation}


It is worth noting that in order to make the generator training eventually converge, here we fix the noise maps $\left\{Z_0, \ldots, Z_N\right\}$ at each scale.

\subsection{Generalization Bound Analysis}

We know from \cite{23} that we can enable better performance with GANs by training the generator and the discriminator using different learning rates. Also, in this case of TTUR training, the generator and discriminator will reach a Nash equilibrium point. This means we can assume that the generator changes slowly enough so that the discriminator can reach the minima approximately independently. In other words, adding adversarial attacks to the discriminator alone allows the discriminator to be trained better without affecting the generator.

\begin{figure}[!t]
	\centering
	\includegraphics[width=3 in]{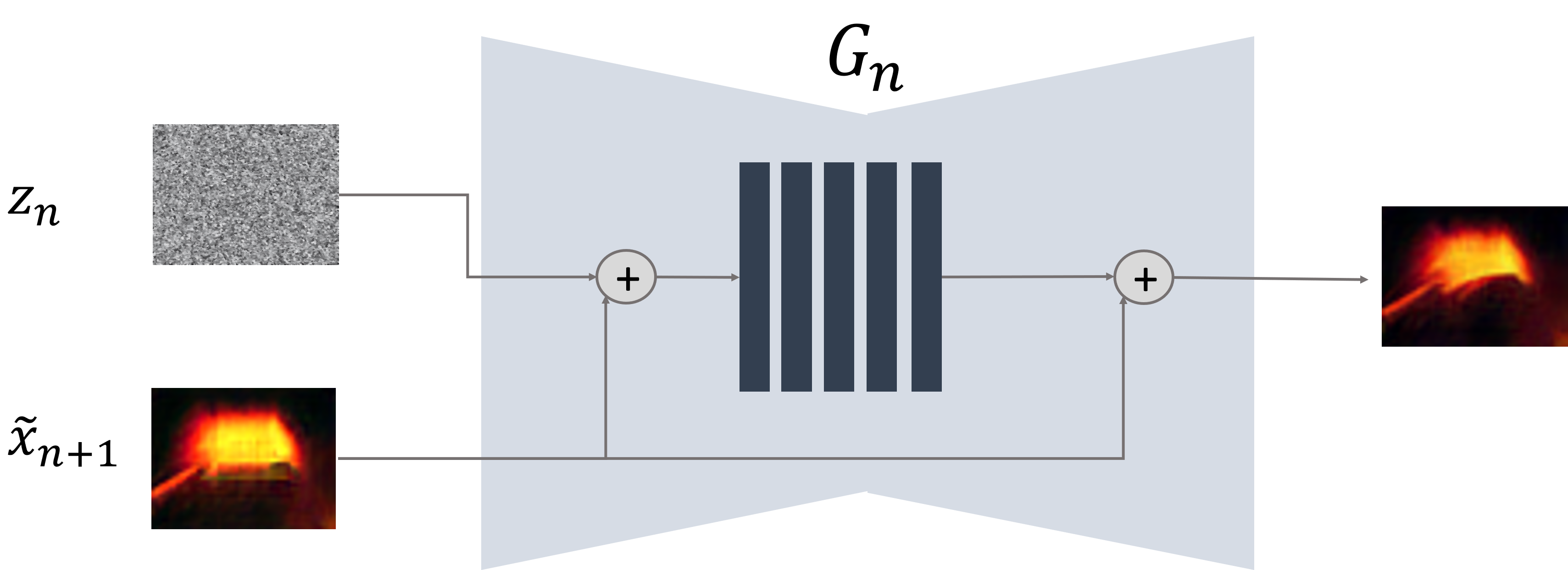}
	\caption{Structure of generators in each scale}
	\label{fig_2}
\end{figure}

On the other hand, we can also treat this problem from the perspective of Rademacher complexity \cite{46}. The traditional generalization error of TTUR is based on restricting the Rademacher complexity, which satisfies Eq. 9.


\begin{equation}
\operatorname{Rad}_n\left(\mathcal{F}_Q\right) \leq 2 Q \sqrt{\frac{2 \ln (2 d)}{n}}
\end{equation}


The Rademacher complexity after the addition of the adversarial attack is shown in Eq. 9. It can be seen that the act of adding adversarial attacks to the discriminator improves the Rademacher complexity of model and hence enable better performance of model.


\begin{equation}
\operatorname{Rad}_n\left(\mathcal{F}_Q\right) \leq 2(1+\varepsilon) Q \sqrt{\frac{2 \ln (2 d)}{n}}
\end{equation}


Considering the way multi-scale GANs are trained, the adversarial training can be seen as an attack injection on the embedding layer of the discriminator for each scale. It is worth noting that the embedding layer in multi-scale GANs is the last layer. In this way, we can transform the problem into a generalized bound problem for adversarial attacks on the last layer of a ReLU-based (which is 1-lipschitz and $\sigma(0)=0$ ) $d$-layer neural network $f_{\mathrm{w}}: \mathcal{X} \rightarrow \mathbf{R}^m$. We denote the weight function of layer $i$ as $\mathbf{w}_i(\cdot)$, and the activation function as $\sigma(\cdot)$. Then for any $f_{\mathbf{w}} \in \mathcal{F}_{\mathrm{nn}}, \mathbf{x} \in \mathcal{X}$, it can be written as $f_{\mathbf{w}}(\mathbf{x})=\mathbf{w}_d \sigma\left(\mathbf{w}_{d-1} \cdots \sigma\left(\mathbf{w}_1 \mathbf{x}\right) \cdots\right)$. In this way the generalization performance of the network can be evaluated using a margin loss as follows.


\begin{equation}
L_\gamma\left(f_{\mathrm{w}}\right)=P\left(f_{\mathrm{w}}(\mathbf{X})[Y] \leq \gamma+\max _{\mathrm{j} \neq Y} f_{\mathrm{w}}(\mathbf{X})[\mathrm{j}]\right)
\end{equation}


Where $\gamma$ is the margin parameter and $f_{\mathrm{w}}(\mathbf{X})[\mathrm{j}]$ is the jth component in the network prediction vector. The implication of the above equation is the difference between the predicted maximum probability value of the $\text{non-}Y$ label and the probability value of the $Y$ label for any given input $\mathbf{X}$. The margin loss allows for a good comparison between empirical risk and true risk.

As mentioned in (3) and (4), adversarial training is a training method that injects a spatially constrained perturbation into the training set during the training process, which can effectively improve the generalization performance of the network. Here we intend to focus on the generalization bounds for FGM and PGD to achieve better adversarial training for multi-scale GANs. The perturbation vector of the FGM can be written as:


\begin{equation}
\epsilon_{\mathbf{w}}^{\mathrm{fgm}}(\mathbf{x})=\underset{\epsilon \in B}{\operatorname{argmax}} \epsilon^T \nabla_{\mathbf{x}} \ell\left(f_{\mathbf{w}}(\mathbf{x}), y\right)
\end{equation}

Instead, the perturbation vector of the PGD can be written as:


\begin{equation}
\begin{gathered}
	\forall 1 \leq t \leq T: \epsilon_{\mathbf{w}}^{\mathrm{pgd}, t+1}(\mathbf{x})=\prod_{\mathcal{B}_{r,\| \cdot \|}(0)}\left\{\epsilon_{\mathbf{w}}^{\mathrm{pgd}, t}(\mathrm{x})+\alpha v_{\mathbf{w}}^{(t)}\right\} \\
	v_{\mathbf{w}}^{(t)}=\underset{\|\epsilon\| \leq 1}{\operatorname{argmax}} \boldsymbol{\epsilon}^T \nabla_{\mathbf{x}} \ell\left(f_{\mathbf{w}}\left(\mathbf{x}+\epsilon_{\mathbf{w}}^{\mathrm{pgd}, t}(\mathbf{x})\right), y\right)
\end{gathered}
\end{equation}

Where $\alpha$ is the step size of each attack and $T$ is the number of steps in the attack. The scope of the attack is restricted by an $r$-bounded norm such that the perturbation vector is in the set $\{\epsilon \in$ $\left.\mathbb{R}^d \mid\|\epsilon\| \leq 1\right\}$. To further evaluate the change in model generalization performance under adversarial training, we define the model margin loss under adversarial training as:


\begin{equation}
\begin{aligned}
	& L_\gamma^{\mathrm{adv}}\left(f_{\mathbf{w}}\right)= \\
	& P\left(f_{\mathbf{w}}\left(\mathbf{X}+\delta_{\mathbf{w}}^{\mathrm{adv}}(\mathbf{X})\right)[Y] \leq \gamma+\max _{\mathrm{j} \neq Y} f_{\mathbf{w}}\left(\mathbf{X}+\delta_{\mathbf{w}}^{\mathrm{adv}}(\mathbf{X})\right)[\mathrm{j}]\right)
\end{aligned}
\end{equation}


After completing the above preliminary definitions, we analyzed such attacks using the PAC-Bayes framework \cite{20}, a theoretical framework proposed by McAllester et al. for evaluating model generalization errors. We eventually obtained generalization bound for two different attacks, and proofs are detailed in the supplementary material.

\textbf{Theorem 1 (Generalization Bound for FGM)} Consider a $d$-layer neural network $f_{\mathbf{w}}: \mathcal{X} \rightarrow \mathbf{R}^m$ with $h$ units in embedding layer with 1-lipschitz activation function $\sigma$ and $\sigma(0)=0$. Suppose that for any $\mathbf{x} \in \boldsymbol{X},\|\mathbf{x}\|_2<B$, and for any $f_{\mathbf{w}}, \kappa \leq\left\|\nabla_{\mathbf{x}} \ell\left(f_{\mathbf{w}}(\mathrm{x}), y\right)\right\|_2$. Let $ \dot{\mathbf{W}}=\prod_{i=1}^d\left\|\mathbf{W}_i\right\|_2$ and $\check{\mathbf{W}} = \sum_{i=1}^d \frac{\left\|\mathrm{W}_i\right\|_F^2}{\left\|\mathrm{W}_i\right\|_2^2}$. Then, we consider an FGM attack with noise power $r$. For any $m, \gamma>0$ with probability $1-\beta$, the following bound holds for the FGM margin loss:


\begin{equation}
	\begin{aligned}
		& L_0^{\mathrm{fgm}}\left(f_{\mathrm{W}}\right) \leq \hat{L}_\gamma^{\mathrm{fgm}}\left(f_{\mathrm{W}}\right) \\
		& +\mathcal{O}\left(\sqrt{\frac{(B+r)^2 d^2 h \log (d h) \boldsymbol{F}_{r, \pi}^{\mathrm{fgm}}\left(f_{\mathrm{w}}\right)+\log \frac{m}{\beta}}{\gamma^2 m}}\right)
	\end{aligned}
\end{equation}


\begin{equation}
	\begin{aligned}
		\boldsymbol{F}_{r, \kappa}^{\mathrm{fgm}}\left(f_{\mathbf{w}}\right)=\left\{\dot{\mathbf{W}}+\frac{r}{\kappa} \dot{\mathbf{W}}(1 / B+\dot{\mathbf{W}})\right\}^2 \check{\mathbf{W}}
	\end{aligned}
\end{equation}

\textbf{Theorem 2 (Generalization Bound for PGD)} Consider $f_{\mathbf{w}}, \mathbf{x}$ and constant $\kappa$ for which the assumptions in Theorem 1 hold. Then, we consider an PGD attack with noise power $r=1, t$ iterations for attack, and step size $\alpha$. For any $m, \gamma>0$ with probability $1-\beta$, the following bound holds for the PGD margin loss:


\begin{equation}
	\begin{aligned}
		& L_0^{\mathrm{pgd}}\left(f_{\mathrm{w}}\right) \leq \hat{L}_\gamma^{\mathrm{pgd}}\left(f_{\mathrm{w}}\right) \\
		& +\mathcal{O}\left(\sqrt{\frac{(B+r)^2 d^2 h \log (d h) F_{r, \kappa, t, \alpha}^{\mathrm{pgd}}\left(f_{\mathrm{w}}\right) +\log \frac{m}{\beta}}{\gamma^2 m}}\right)
	\end{aligned}
\end{equation}


\begin{equation}
	\begin{aligned}
		& \boldsymbol{F}_{r, \kappa, t, \alpha}^{\mathrm{pgd}}\left(f_{\mathbf{w}}\right)= \\
		& \left\{\frac{1-(2 \alpha / \kappa)^t\times\overline{l i p}\left(\nabla \ell \circ f_{\mathbf{w}}\right)^t}{\kappa-2 \alpha \times \overline{l i p}\left(\nabla \ell \circ f_{\mathbf{w}}\right)}\right\}^2 \dot{\mathbf{W}}(1+\dot{\mathbf{W}})\check{\mathbf{W}}
	\end{aligned}
\end{equation}

Where $ \overline{l i p}\left(\nabla \ell \circ f_{\mathbf{w}}\right) =\dot{\mathbf{W}} \sum_{i=1}^d \prod_{j=1}^i\left\|\mathbf{W}_j\right\|_2$, which provides an upper-bound on the Lipschitz constant of the gradient $\nabla_{\mathbf{x}} \ell\left(f_{\mathbf{w}}(\mathbf{x}), y\right)$.

As we only need to attack the embedding layer in multi-scale GANs, the perturbation bound and gradient perturbation bound is tighter than before \cite{42,47}. So that in this way, we obtain a more reliable generalization performance evaluation metric for multi-scale GANs.

Besides, comparing the generalization bound under the two attacks, we can find that when $(2 \alpha / \kappa)^t \overline{l i p}\left(\nabla \ell \circ f_{\dot{\mathrm{w}}}\right)^t<1$, for any number of steps of PGD, the margin-based generalization bound will grow the FGM generalization error bound by factor $\overline{l i p}\left(\nabla \ell \circ f_{\mathbf{w}}\right)(2 \alpha / \kappa)$. This phenomenon suggests that during training, when $\overline{l i p}\left(\nabla \ell \circ f_{\mathrm{w}}\right)<\kappa /(2 \alpha)$, we need to change the attack method, which inspire us to propose the final adaptive adversarial training approach.

\subsection{Adaptive Adversarial Training Method}

In order to improve the generalization of discriminator, we need to vary the attack methods as generalization bound changes during the training process. Our adaptive adversarial training method was created as following. Firstly, we calculate the forward loss for each input $\mathrm{x}$. Then we compute the backpropagation gradient and back it up. Afterwards, we evaluate $W_{\boldsymbol{\sigma}}$. If $W_{\boldsymbol{\sigma}}<\kappa /(2 \alpha)$, we use the PGD attack approach. If $W_\sigma>\kappa /(2 \alpha)$, the FGM attack approach is adopted. It is worth noting that $\overline{l i p}\left(\nabla \ell \circ f_{\mathbf{w}}\right)$ cannot be computed directly, so here we use the largest singular value of the weight matrix $W_\sigma$ as the evaluation metric for changing the attack method.

\IncMargin{1em}
\begin{algorithm} 
	 \SetKwInOut{Input}{input}\SetKwInOut{Output}{output}
	
	\Input{Training dataset $\mathcal{X}$, Initial model $f_w(\mathrm{x})$} 
	\Output{Final model $F_{\mathrm{w}}(\mathbf{x})$}
	\BlankLine 
	
	\For{$\mathrm{x} \in \mathcal{X}:$}{ 
		\emph{Calculate the forward loss of $\mathbf{x}$}\; 
		\emph{Back propagate to get $\boldsymbol{\nabla}_{\mathbf{x}} \ell\left(f_{\mathbf{w}}(\mathbf{x}), y\right)$ }\; 
		\emph{Back up}\;
		\If{$W_\sigma<\kappa /(2 \alpha)$}{
			$$
			\epsilon_{\mathrm{w}}^{\mathrm{pgd}, 1}(\mathrm{x})=\prod_{\mathcal{B}_{r,\|\cdot \|}(0)}\left\{\alpha v_{\mathrm{w}}^{(0)}\right\}
			$$
			
			Where $v_{\mathrm{w}}^{(0)}=\underset{\|\epsilon\| \leq 1}{\operatorname{argmax}} \epsilon^T \nabla_{\mathrm{x}} \ell\left(f_{\mathrm{w}}(\mathbf{x}), y\right)$
			
			\textbf{Update} {$f_{\mathrm{w}}(\mathbf{x}) \leftarrow f_{\mathrm{w}}\left(\mathrm{x}+\epsilon_{\mathrm{w}}^{\mathrm{pgd}, \mathrm{1}}(\mathrm{x})\right)$};
			
			\BlankLine 
			\For{$1<t \leq T$}{ 
				$$
				\epsilon_{\mathrm{w}}^{\mathrm{pgd}, t}(\mathbf{x})=\prod_{\mathcal{B}_{r,\|\cdot\|}(0)}\left\{\epsilon_{\mathrm{w}}^{\mathrm{pgd}, t-1}(\mathbf{x})+\alpha v_{\mathrm{w}}^{(t-1)}\right\}
				$$
				
				Where $v_{\mathrm{w}}^{(t-1)}=\underset{\|\epsilon\| \leq 1}{\operatorname{argmax}} \epsilon^T \nabla_{\mathbf{x}} \ell\left(f_{\mathrm{w}}\left(\mathrm{x}+\epsilon_{\mathrm{w}}^{\mathrm{pgd}, t-1}(\mathrm{x})\right), y\right)$ 
				
				\textbf{Update} {$f_{\mathrm{w}}(\mathbf{x}) \leftarrow f_{\mathrm{w}}\left(\mathrm{x}+\epsilon_{\mathrm{w}}^{\mathrm{pgd}, t}(\mathrm{x})\right)$;}
			}}
		\Else{
			$$
			\epsilon_{\mathrm{w}}^{\mathrm{fgm}}(\mathbf{x})=\underset{\epsilon \in B^{\mathrm{fgm}}}{\operatorname{argmax}} \epsilon^T \nabla_{\mathrm{x}} \ell\left(f_{\mathrm{w}}(\mathbf{x}), y\right)
			$$
			
			\textbf{Update} $f_{\mathrm{w}}(\mathrm{x}) \leftarrow f_{\mathrm{w}}\left(\mathrm{x}+\epsilon_{\mathrm{w}}^{\mathrm{fgm}}(\mathrm{x})\right)$}
		} 
	{$F_{\mathrm{w}}(\mathrm{x}) \leftarrow \boldsymbol{f}_{\mathrm{w}}(\mathrm{x})$}
	\caption{Adaptive Adversarial Training}
	\label{alg1} 
\end{algorithm}
\DecMargin{1em}

\section{Evaluation and Experiment}
\subsection{Experiment Setup}

We used Nvidia RTX 2080Ti GPU for model training and inference. Similar to SinGAN \cite{17}, we used the Berkeley Segmentation Database (BSD) \cite{48}, Places \cite{49} and some image data from the web for model performance evaluation. As for the hyperparameters of model, we set the coarsest scale to 25px and the finest scale to 250px in order to compare the models fairly. In order to make the scale factor\ r\ closely to 4/3, the scale of the different images is varied by layer N. 

To validate the gain of the proposed adaptive adversarial training method for multi-scale GANs, the model is tested in a variety of image processing tasks. Firstly, a series of paired image to single image experiments were conducted to verify the style fusion capability of the improved multi-scale GANs. We used SIFID to evaluate multi-scale GANs with different training methods on the paint-to-image task, followed by an AMT study on the traditional art style transfer task. Afterwards, we carried out classical image super-resolution restoration experiments to evaluate the generative power of the improved multi-scale GANs on the single image to single image tasks using NIQE and RMSE metrics.

\begin{figure}[!t]
	\centering
	\includegraphics[width=3.2 in]{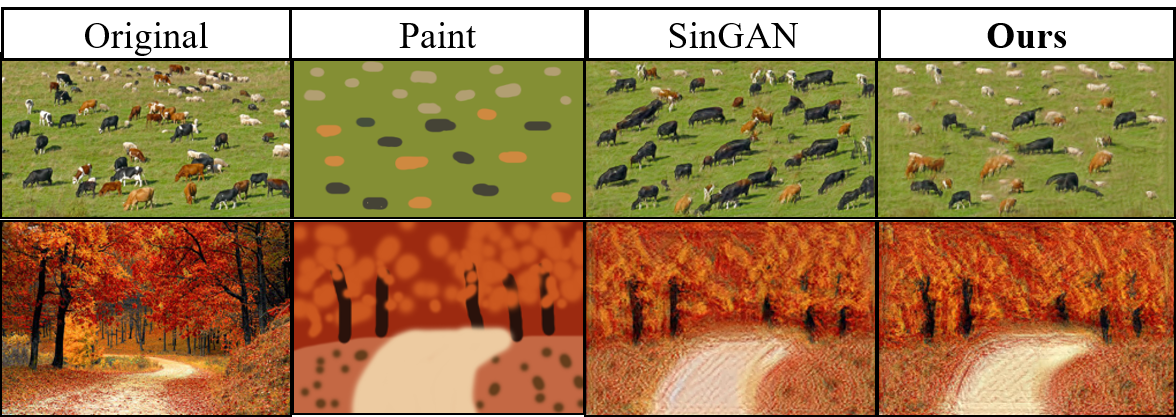}
	\caption{Paint to image on complex textured images and simple textured images}
	\label{fig_3}
\end{figure}

\begin{figure}[!t]
	\centering
	\includegraphics[width=3.2 in]{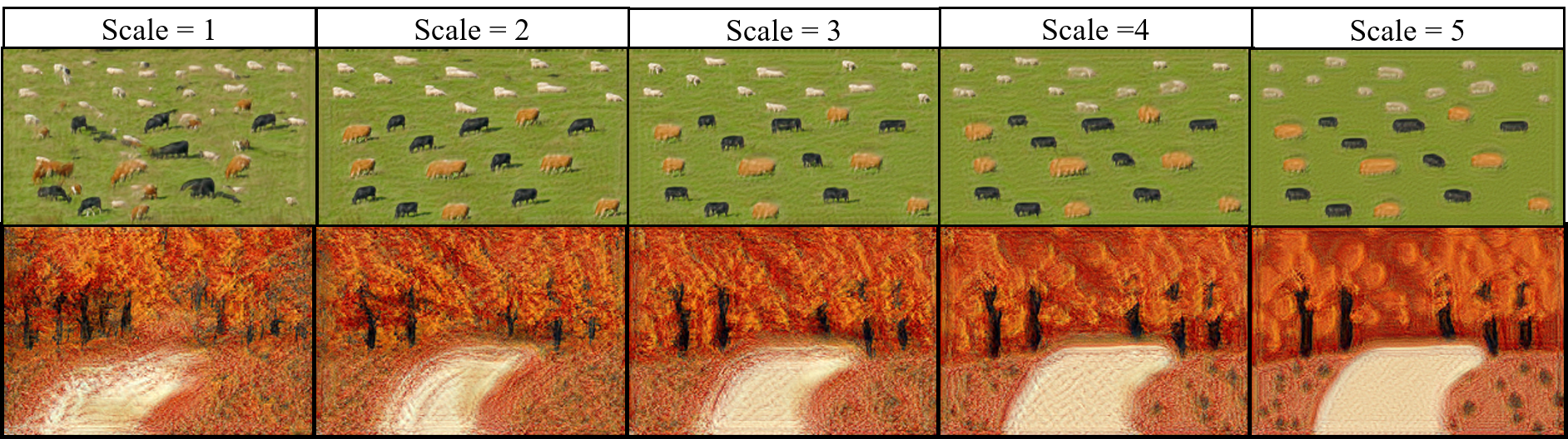}
	\caption{Generating results for paint to image tasks with different scale injections}
	\label{fig_4}
\end{figure}


\subsection{Paint to Image}

\begin{table*}[h]
	\centering
	\setlength{\tabcolsep}{5.5 mm}
	\caption{SIFID for simple texture images in the paint to image task}
	\begin{tabular}{lclclcl} 
		\toprule 
		SIFID↓ (/1e-5) &  Scale=1 & Scale=2 & Scale=3 & Scale=4 & Scale=5 \\
		\midrule 
		\text { SinGAN } & \textbf{7.811} & 9.201 & 12.880 & 9.562 & 8.119 \\
		\text { SinGAN+TTUR } & \textbf{6.995} & 8.834 & 12.848 & 9.576 & 8.207 \\
		\text { SinGAN+TTUR+FGM } & \textbf{5.707} & 7.347 & 12.447 & 9.695 & 8.801 \\
		\text { SinGAN+TTUR+PGD } & \textbf{4.434} & 5.546 & 9.556 & 6.996 & 6.450 \\
		\text { SinGAN+TTUR+ours } & \textbf{3.936} & \textbf{5.375} & \textbf{9.544} & \textbf{6.936} & \textbf{6.381} \\
		\bottomrule 
	\end{tabular}
	\label{tab1} 
\end{table*}

\begin{table*}[h]
	\centering
	\setlength{\tabcolsep}{5.5 mm}
	\caption{SIFID for complex texture images in the paint to image task}
	\begin{tabular}{lclclcl} 
		\toprule 
		SIFID↓ (/1e-5) &  Scale=1 & Scale=2 & Scale=3 & Scale=4 & Scale=5 \\
		\midrule 
		\text { SinGAN } & \textbf{7.811} & 9.201 & 12.880 & 9.562 & 8.119 \\
		\text { SinGAN+TTUR } & \textbf{6.995} & 8.834 & 12.848 & 9.576 & 8.207 \\
		\text { SinGAN+TTUR+FGM } & \textbf{5.707} & 7.347 & 12.447 & 9.695 & 8.801 \\
		\text { SinGAN+TTUR+PGD } & \textbf{4.434} & 5.546 & 9.556 & 6.996 & 6.450 \\
		\textbf { SinGAN+TTUR+Ours } & \textbf{3.936} & \textbf{5.375} & \textbf{9.544} & \textbf{6.936} & \textbf{6.381} \\
		\bottomrule 
	\end{tabular}
	\label{tab2} 
\end{table*}

The paint to image task involves combining a simple graffiti painting with a realistic reference image to produce a new photograph with the structure of the graffiti painting and the realistic style of the reference photograph. We use the reference image as training data in our experiments and generate the final new photograph by learning the visual features from it. More specifically, we replace the coarser scale noise map in the trained network with a graffiti painting, and the final output of the network is the new photograph.

As shown in Figure 3, we directly use the original image and the graffiti image provided in SinGAN for training, after which the graffiti image is injected for each scale and the final best generated new image is obtained. A side-by-side comparison of the generated new images shows that the proposed method works better for images with simple texture features, but for the original images with more complex texture features, it is difficult to judge which model produces better.

To quantify the performance of the model, we use SIFID for the evaluation of paint to image task. SIFID test is a method that uses the Inception v3 \cite{50} for image feature extraction and measure the similarity of the two images based on these features. It calculates the discrete Fréchet distance of the internal distribution of the features as SIFID, and smaller SIFID indicate higher image quality.

We tested injected graffiti image at different scales for the original SinGAN, TTUR-based SinGAN, and TTUR-based SinGAN with different attacks, respectively. It is worth noting that different images here tend to have different optimal injection scales, so here we compare the minimum value of SIFID for different scales of injection. The images generated at different scales are shown in Figure 4. It can be seen that the structure of the resulting new photo is consistent with graffiti painting, but with a stylistic bias towards a real photo. The final calculated SIFID are shown in Table 1 and Table 2. By comparing the SIFID of the new pictures generated after the injection of real images performed at different scales, we find that the adaptive attack method proposed in this paper has considerable advantages, both in terms of the best SIFID and the SIFID of the graffiti paintings injected at each different scale. From the SIFID results of the complex texture images we can see that the final best image generated by the method proposed in this paper has a 100\% drop in SIFID compared to the original SinGAN, while under the complex texture image the SIFID proposed in this paper has a 200\% drop.

\subsection{Art Style Transfer}

Style Transfer is a classical image manipulation task based on GANs, which obtains a new picture with the artist's style by implanting the artist's painting style into the expected content. Figure 5 shows an example of a set of style and content images. We tested the original SinGAN, the TTUR-based SinGAN and the TTUR-based SinGAN under different attacks with this example. The results are shown in Figure 6. It can be seen that the SinGAN using the adaptive attack method proposed in this paper successfully blends the content and the artist's style, expecting that the overall structure of the content and the texture and high frequency information matching the artist's drawing are well preserved. Visually, the SinGAN incorporating the adaptive attack produces the best images.

For quantitative comparison of effects, the AMT Study was used to evaluate the effectiveness of the model. We created 20 samples using images collected on the web and presented them to 30 persons with experience in computer vision, asking them to choose the better result. The final test results are shown in Table 3. SinGAN with our adaptive training has a significantly higher preference rate than the original one, while the difference is quite large, at 63.67\%.

\begin{figure}[!t]
	\centering
	\includegraphics[width=3.2 in]{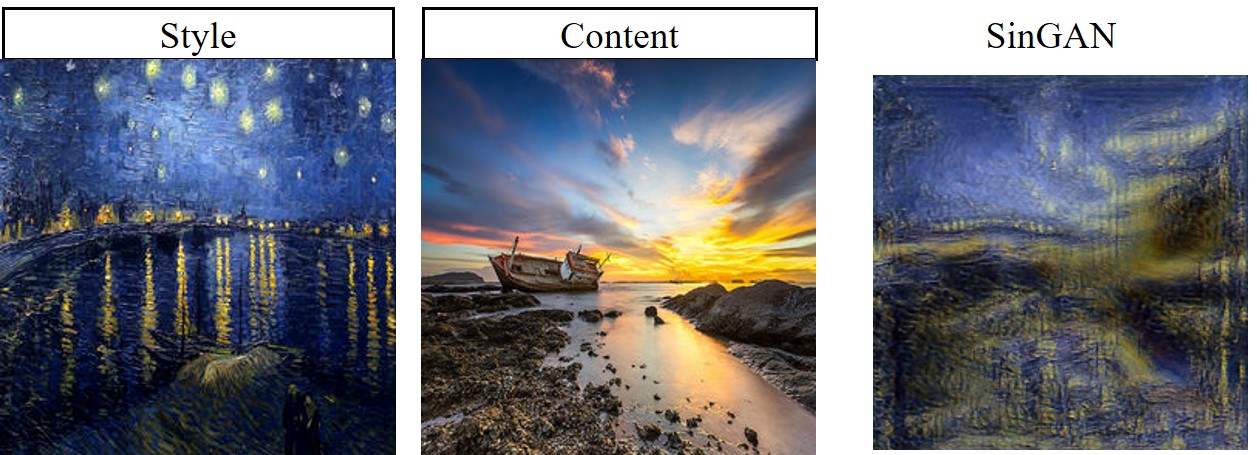}
	\caption{Style and content images}
	\label{fig_5}
\end{figure}

\begin{figure}[!t]
	\centering
	\includegraphics[width=3.2 in]{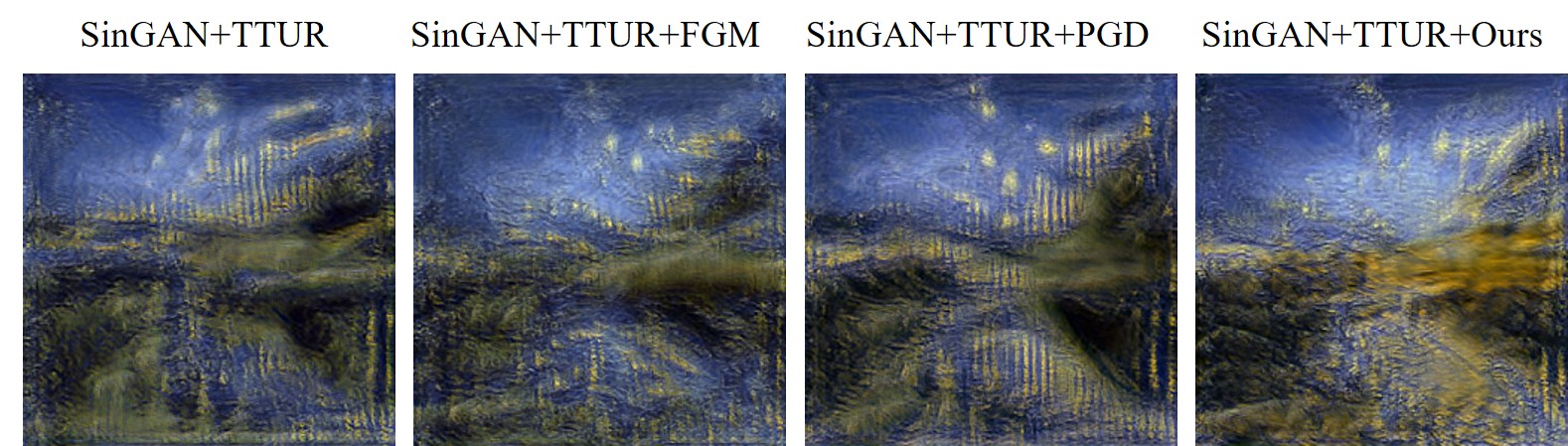}
	\centering\caption{Art style transfer result}
	\label{fig_6}
\end{figure}

\begin{table}[htbp]
	\centering
	\setlength{\tabcolsep}{5.5 mm}
	\caption{AMT study result for art style transfer}
	\begin{tabular}{lclclcl} 
		\toprule 
		 &  Art Style Transfer \\
		\midrule 
		\text { SinGAN } & 5.00\% (30/600) \\
		\text { SinGAN+TTUR } & 7.67\% (46/600) \\
		\text { SinGAN+TTUR+FGM } & 10.17\% (61/600) \\
		\text { SinGAN+TTUR+PGD } & 13.50\% (81/600) \\
		\textbf { SinGAN+TTUR+Ours } & \textbf{63.67\% (382/600)}  \\
		\bottomrule 
	\end{tabular}
	\label{tab3} 
\end{table}

\subsection{Super-Resolution Restoration}

The image super-resolution restoration with multi-scale GANs are slightly different from the two tasks described above. In the image super-resolution restoration task, the model needs to refine the detailed textures while fine-tuning the overall image structure. To achieve these, we up-sampled the low-resolution image appropriately and mix in some noise before feeding it into $\mathrm{G}_0$, and at the same time, we set the reconstruction loss weight $\alpha=100$ and pyramid scale factor $r=\sqrt[k]{s}$, where $\mathrm{k}\in$ N. Figure 7 shows an example of image super-resolution restoration using different training methods. Compared to other training methods, the adaptive training method proposed in this paper ends up with much less color distortion than the other training methods.

Further, we compared the super-resolution restoration result of the model under the training approach proposed in this paper with external training methods on large-scale datasets (ESRGAN \cite{8}, DBPN \cite{51}) and the self-similarity based super-resolution method (SelfExSR \cite{52}). As shown in Figure 8, the visual quality of our single-image based reconstruction of high-resolution images is high and our method is comparable to SRGAN and DBPN trained on large-scale external datasets.

To evaluate the restoration quality, we used the BSD100 dataset for 4x super-resolution restoration and tested the distortion and perceived quality of the images restored by the different models using metrics such as RMSE and NIQE, respectively. RMSE and NIQE are two conflicting super-resolution criteria, but both give an indication of the quality of the images \cite{53}. From the results in Table 4 we can see that the proposed method significantly reduces the distortion and improves the perceptual quality of the images, with a 60\% reduction in RMSE and a 100\% reduction in NIQE. It is worth noting that, the image perception quality of Multi-Scale GAN with our adaptive attack even leaps to the best of all models.


\begin{table}[htbp]
	\centering
	\setlength{\tabcolsep}{5 mm}
	\caption{Image quality evaluation for super-resolution restoration}
	\begin{tabular}{lclclcl} 
		\toprule 
		&  RMSE↓ &  NIQE↓\\
		\midrule 
		\text { SelfExSR \cite{52} } & 508.074 & 6.159\\
		\text { ESRGAN \cite{8} } & 417.905 & 3.642\\
		\text { DBPN \cite{51} } & \textbf{263.069} & 6.610\\
		\text { SinGAN } & 1098.043 & 4.182\\
		\text { SinGAN+TTUR } & 1047.378 & 4.314\\
		\text { SinGAN+TTUR+FGM } & 1431.025 & 4.094\\
		\text { SinGAN+TTUR+PGD } & 1017.990 & 6.006\\
		\textbf { SinGAN+TTUR+Ours } & \textbf{424.734}  & \textbf{2.165}\\
		\bottomrule 
	\end{tabular}
	\label{tab4} 
\end{table}

\begin{figure}[!t]
	\centering
	\includegraphics[width=3.2 in]{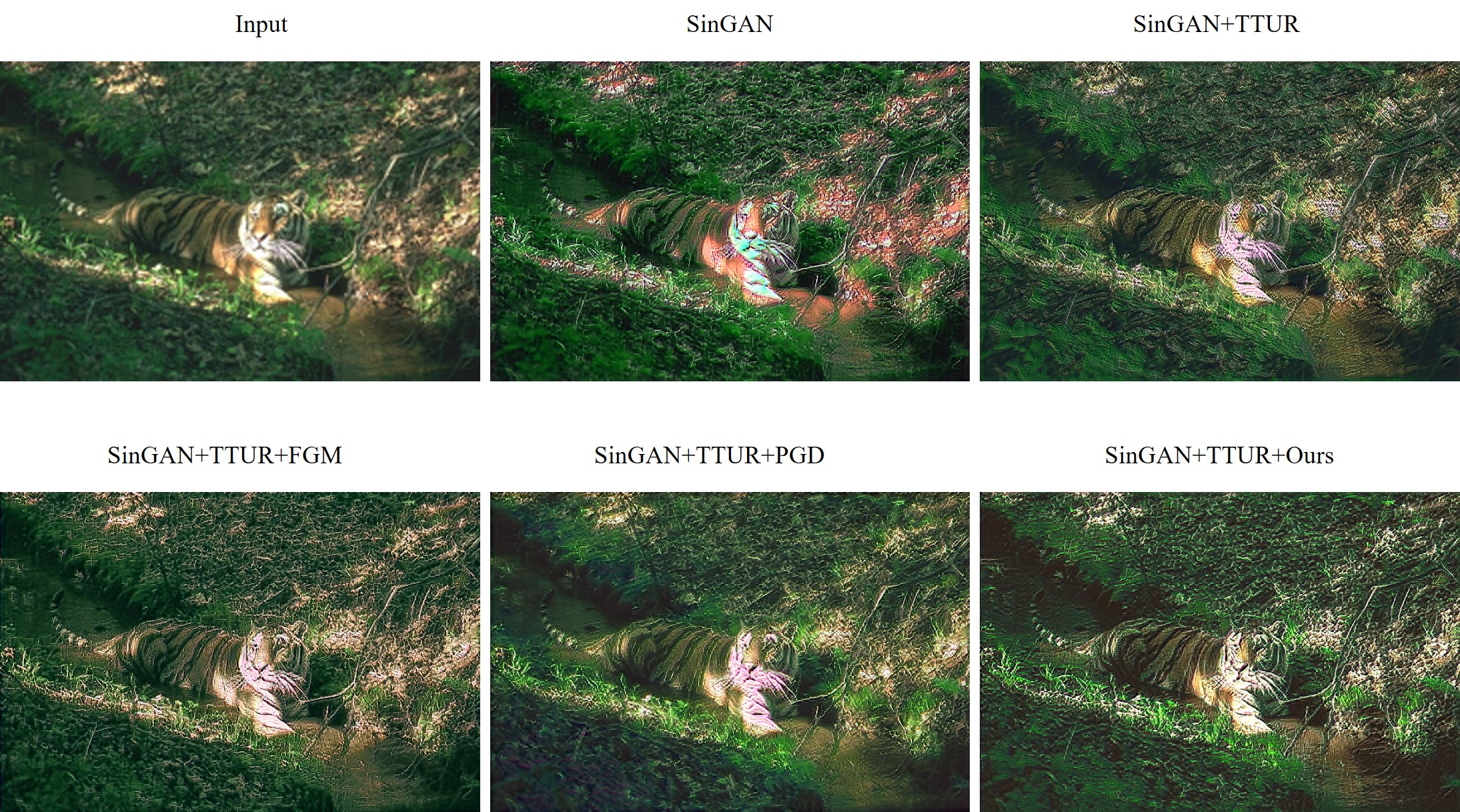}
	\caption{Super-resolution restoration results for different training methods}
	\label{fig_7}
\end{figure}

\begin{figure}[!t]
	\centering
	\includegraphics[width=3.2 in]{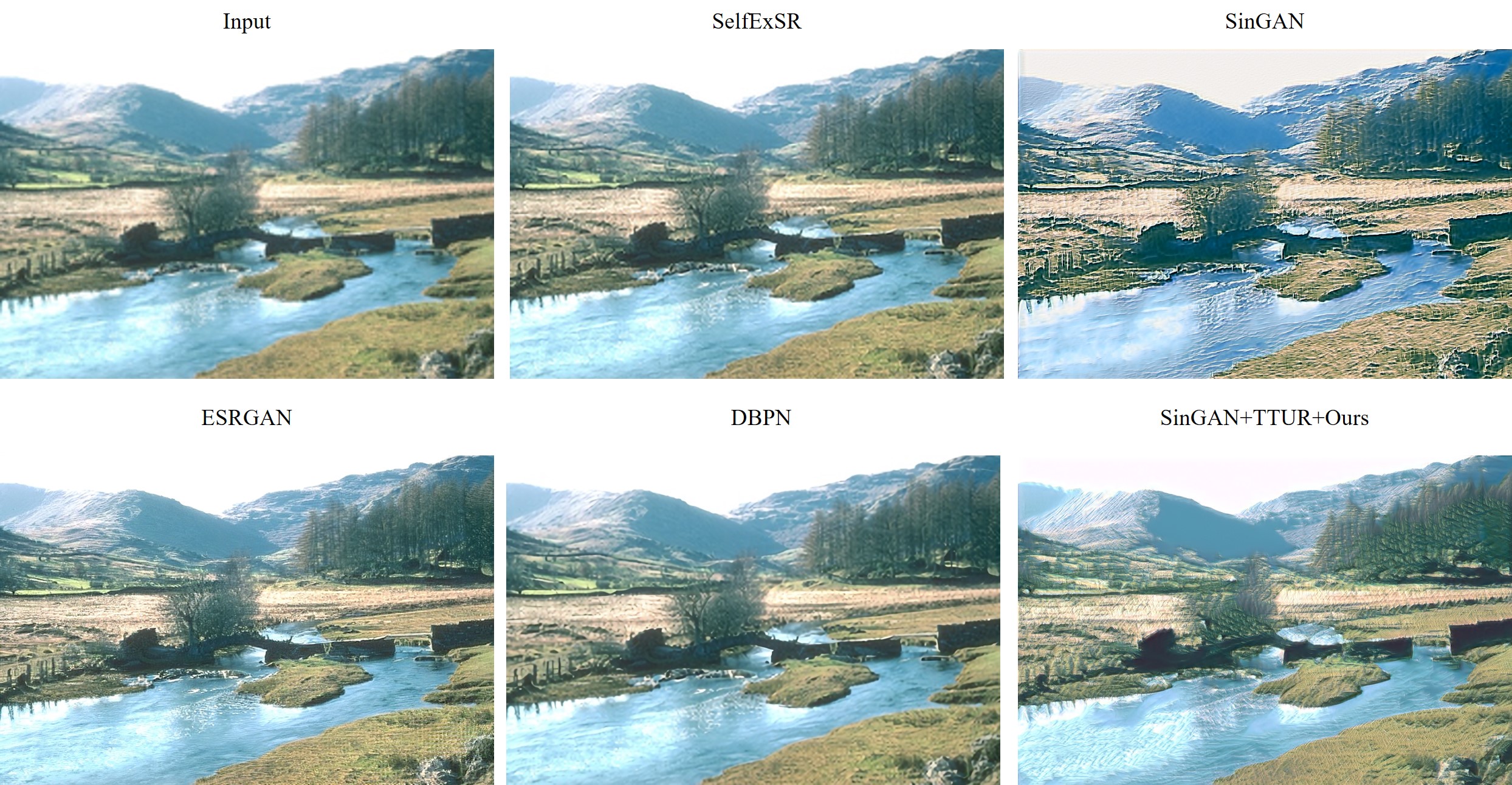}
	\caption{Super-resolution restoration results for different models}
	\label{fig_8}
\end{figure}


\section{Conclusion}
We present an adaptive adversarial training method for multi-scale GANs. First, we use TTUR to optimize the multi-scale GAN, while this operation allows the GAN to achieve Nash equilibrium during training. Simplifying the generalization bound analysis for multi-scale GANs based on the Nash equilibrium, we propose an adaptive adversarial training method which can greatly improve the quality of multi-scale GAN-generated images. The effectiveness of the proposed attack approach is evaluated on different tasks, and results show that multi-scale GANs based on our adaptive adversarial attack approach have significant improvement on all tasks we have tried. It is worth noting that our analysis for a particular model can also be an inspiration for later training optimization research, and we will continue to focus on effective optimization under the theoretical analysis of different models in the future.

\bibliography{bib.bib}
\bibliographystyle{unsrt}

\vspace{15mm}

\begin{wrapfigure}{l}{25mm} 
	\vspace{-5mm}
	\includegraphics[height=1.5in,keepaspectratio]{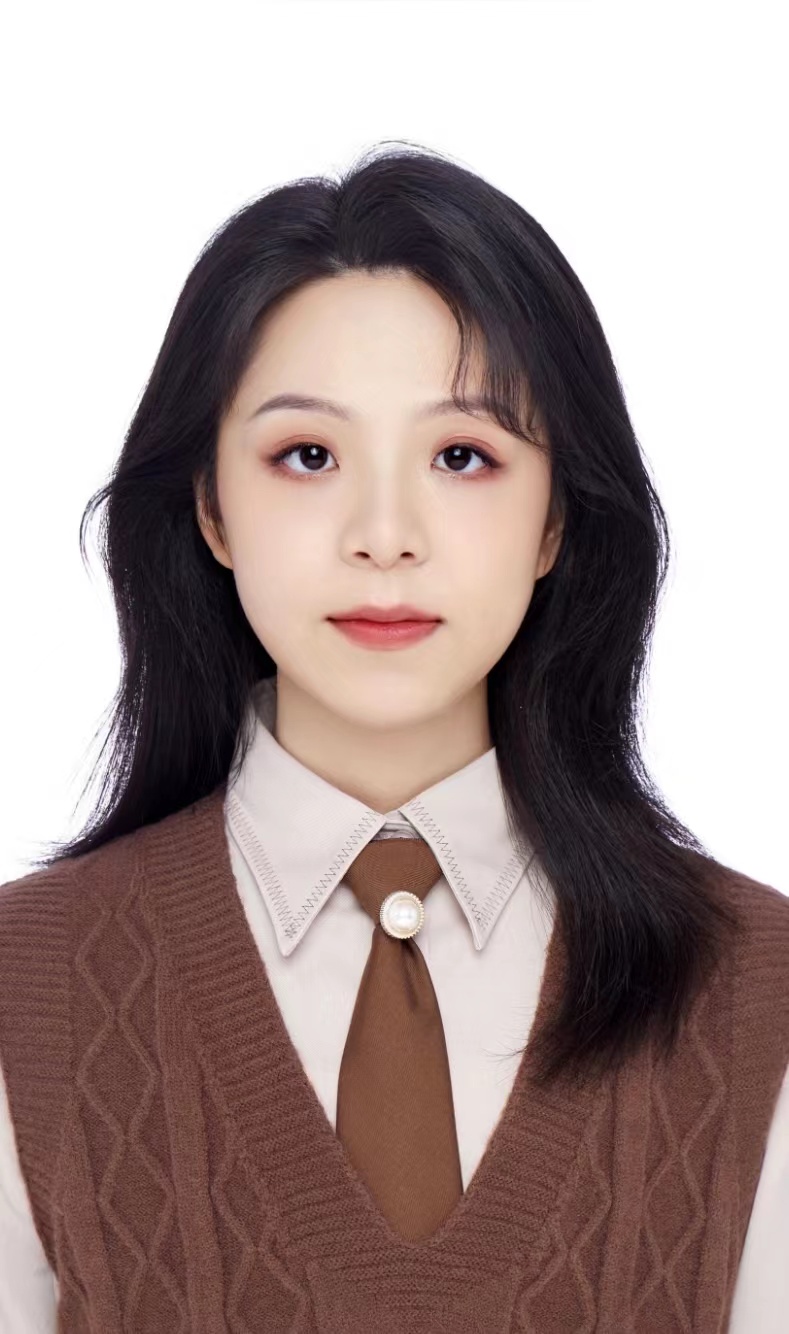}
\end{wrapfigure}\par
\textbf{Jing Tang} is currently pursuing her Ph.D. degree in Mechanical Engineering at Huazhong University of Science and Technology(HUST), Wuhan, China. Her research is mainly related to robot dexterous manipulation, neural network architecture research. Her current research interests are focused on robot grasping based on machine learning methods.\par

\vspace{35mm}

\begin{wrapfigure}{l}{25mm} 
	\vspace{3mm}
	\includegraphics[height=1.4 in,keepaspectratio]{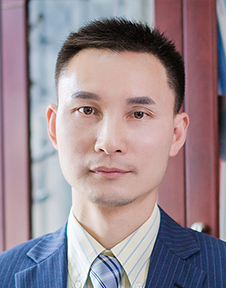}
\end{wrapfigure}\par
\textbf{Bo Tao} received the Ph.D. degree in mechanical engineering from the Huazhong University of Science and Technology (HUST), Wuhan, China, in 2007.He is currently a Professor and a Doctoral Supervisor with the School of Mechanical Science and Engineering, HUST. He has authored more than 80 articles in international journals. His current research interests include intelligent. \par

\vspace{15mm}


\vspace{40mm}

\begin{wrapfigure}{l}{25mm} 
	\vspace{-2mm}
	\includegraphics[height=1.4 in,keepaspectratio]{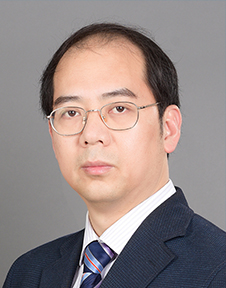}
\end{wrapfigure} \par
\textbf{Zhouping Yin}	received the Ph.D. degree in mechanical engineering from the Huazhong University of Science and Technology (HUST), Wuhan, China, in 2000.He is currently a Professor with the School of Mechanical Science and Engineering, HUST. He has authored more than 100 SCI/EI articles and three monographs. His research interests involve digital intelligent manufacturing technology and electronic manufacturing technology and equipment. \par

\vspace{-105mm}

\begin{wrapfigure}{l}{25mm} 
	\vspace{-2mm}
	\includegraphics[height=1.4 in,keepaspectratio]{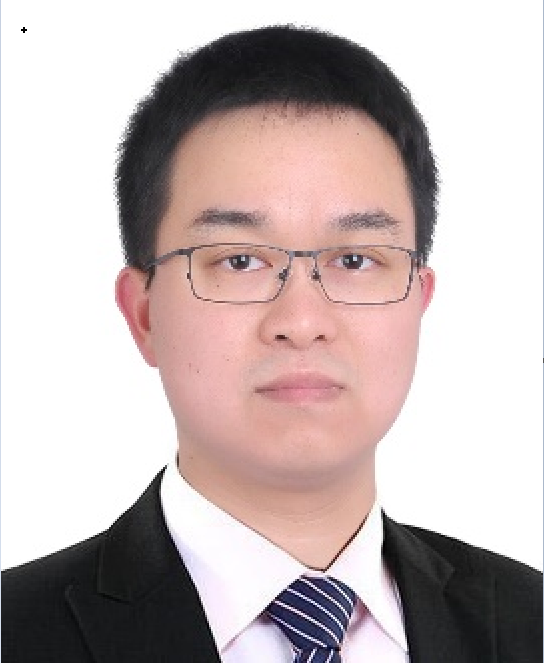}
\end{wrapfigure}\par
\textbf{Zeyu Gong}  received the B.S. and Ph.D. degrees in mechanical engineering from Huazhong University of Science and Technology in 2012 and 2018 respectively. He is currently a Lecturer with the Department of Mechanical Science and Engineering, HUST. His research interests include robot state estimation, visual servoing and climbing robot. \par




%
%
%
%
%
%
%
%
%
%
%

\end{document}